\documentclass[conference]{IEEEtran}
\IEEEoverridecommandlockouts
\usepackage{cite}
\usepackage{amsmath,amssymb,amsfonts}
\usepackage{algorithmic}
\usepackage{graphicx}
\usepackage{textcomp}
\usepackage{xcolor}
\usepackage{mathrsfs}
\usepackage{url}
\def\BibTeX{{\rm B\kern-.05em{\sc i\kern-.025em b}\kern-.08em
    T\kern-.1667em\lower.7ex\hbox{E}\kern-.125emX}}

\newcommand{\PREPRINTYEAR}{2024}
\newcommand{\PUBLISHEDIN}{IEEE International Conference on Control, Automation, Robotics, and Vision }
\newcommand{\DOI}{10.1109/ICARCV63323.2024.10821603} 

\usepackage[placement=top,vshift=-2,firstpage=true]{background}
\SetBgScale{1.0}
\SetBgContents{\parbox{1.00\textwidth}{\small \begin{center} The manuscript was published in \PUBLISHEDIN under DOI: \DOI\end{center} \vspace{-0.2cm}  \copyright{}~\PREPRINTYEAR~IEEE.~Personal use of this material is permitted. Permission from IEEE must be obtained for all other uses, in any current or future media, including reprinting/republishing this material for advertising or promotional purposes, creating new collective works, for resale or redistribution to servers or lists, or reuse of any copyrighted component of this work in other works.} }
\SetBgColor{black}
\SetBgAngle{0}
\SetBgOpacity{1.0}

\begin{document}

\title{\rule{0mm}{10mm}On rapid parallel  tuning of controllers of a swarm of MAVs -- distribution strategies of the  updated gains \\
\thanks{This work has been funded by Poznan
University of Technology under project 0214/SBAD/0247 and by the Czech Science Foundation (GA\v{C}R) under research project no.~23-07517S, by the Czech Science Foundation (GA\v{C}R) under research project no.~22-24425S and by the European Union under the project Robotics and advanced industrial production (reg. no. CZ.02.01.01/00/22\_008/0004590).

The source code of the EQL method is available at https://github.com/AppliedControlTechniques/MAV\_parallel\_tuning\_EQL, and a short movie presenting parallel tuning on YouTube https://youtu.be/kouVJeRgWjI.
}
}

\author{\IEEEauthorblockN{Dariusz Horla}
\IEEEauthorblockA{\textit{Faculty of Automation, Robotics and Electrical Engineering} \\
\textit{Institute of Robotics and Machine Intelligence}\\
\textit{Poznan University of Technology}\\
Poznan, Poland \\
dariusz.horla@put.poznan.pl}
\and
\IEEEauthorblockN{Wojciech Giernacki}
\IEEEauthorblockA{\textit{Faculty of Automation, Robotics and Electrical Engineering} \\
\textit{Institute of Robotics and Machine Intelligence}\\
\textit{Poznan University of Technology}\\
Poznan, Poland \\
wojciech.giernacki@put.poznan.pl}
\and
\IEEEauthorblockN{V\'{i}t Kr\'{a}tk\'{y}}
\IEEEauthorblockA{\textit{Department of Cybernetics, Faculty of Electrical Engineering} \\
\textit{Czech Technical University in Prague}\\
Prague, Czechia \\
vit.kratky@fel.cvut.cz}
\and
\IEEEauthorblockN{Petr \v{S}tibinger}
\IEEEauthorblockA{\textit{Department of Cybernetics, Faculty of Electrical Engineering} \\
\textit{Czech Technical University in Prague}\\
Prague, Czechia \\
petr.stibinger@fel.cvut.cz}
\and
\IEEEauthorblockN{Tom\'{a}\v{s} B\'{a}\v{c}a}
\IEEEauthorblockA{\textit{Department of Cybernetics, Faculty of Electrical Engineering} \\
\textit{Czech Technical University in Prague}\\
Prague, Czechia \\
tomas.baca@fel.cvut.cz}
\and
\IEEEauthorblockN{Martin Saska}
\IEEEauthorblockA{\textit{Department of Cybernetics, Faculty of Electrical Engineering} \\
\textit{Czech Technical University in Prague}\\
Prague, Czechia \\
martin.saska@fel.cvut.cz}
}

\maketitle

\begin{abstract}
In this paper, we present a reliable, scalable, time deterministic, model-free procedure to tune swarms of Micro Aerial Vehicles (MAVs) using basic sensory data. Two approaches to taking advantage of parallel tuning are presented. First, the tuning with averaging of the results on the basis of performance indices reported from the swarm with identical gains to decrease the negative effect of the noise in the measurements. Second, the tuning with parallel testing of varying set of gains across the swarm to reduce the tuning time. The presented methods were evaluated both in simulation and real-world experiments. The achieved results show the ability of the proposed approach to improve the results of the tuning while decreasing the tuning time, ensuring at the same time a reliable tuning mechanism. 
\end{abstract}

\begin{IEEEkeywords}
tuning, optimization, tracking performance, swarm, MAVs
\end{IEEEkeywords}

\section{Introduction}
In the paper, we show how to rapidly tune controller gains using a parallel tuning method scalable from a pair, to a swarm of Micro Aerial Vehicles (MAVs), with an approachable iterative-based and fully deterministic in time algorithm. The method increases both the efficiency, reliability, as well as decreases tuning time expense. The outline of the algorithm for a single MAV has been presented in \cite{b5} for a Fibonacci-based algorithm. Next, it has been extended to a family of zero-order algorithms in \cite{b3}, and implemented to offer a rapid tuning, as reported in \cite{b4}. In this paper, an approach for deployment of these tuning methods to enable parallel tuning is presented, to improve the performance and decrease the time expenses.

There are multiple examples showcasing the parallel actions are advantageous in various fields. The list can include, e.g., cases in genome sequencing \cite{b6} (analyzing different sections of a genome simultaneously to significantly reduce the time required to sequence an entire genome), in climate modelling \cite{b7} (running multiple simulations in parallel to compare different models quicker, to improve reliability of climate predictions), and in linear algebra \cite{b8} (sparse/large matrices divided into  sub-matrices,  multiplied concurrently, to reduce computation time). From the field of engineering: in distributing the training process across multiple GPUs for deep learning models, in IT sciences \cite{b9} (multiple client requests handled  concurrently to improve response time), in assembly tasks (different components of a product assembled simultaneously at various stations to speed-up the overall process), or in quality control (quality checksin parallel on different production lines to promptly detect defects).

As in the case of bio-inspired optimization algorithms, examples can be found in nature. Ants use parallel action in their foraging behavior \cite{b10} (simultaneous food search to increase  the likelihood of finding resources quickly), or in nest building (digging, transporting materials, and defending the colony simultaneously to speed up the nest-building process). Also in plant growth, multiple leaves on a plant engage in photosynthesis concurrently (to maximize energy production)  or during roots grow (efficient exploration and exploiting of soil resources). Bees visit numerous flowers simultaneously (to enhance pollination process), or collect nectar, produce honey to ensure a steady supply to maintain hive. Birds migrate in flocks (parallel actions used to navigate), fish in a school move in a coordinated manner \cite{b11} (to evade predators efficiently), or wolves hunt in packs where different members take on roles such as chasing, flanking, and ambushing prey simultaneously. 
Parallel actions in nature enhance efficiency, adaptability, and survival across various species. By engaging in parallel, organisms optimize their use of resources, ensure reproductive success, and maintain the stability of their ecosystems.

Finally, in the engineering context, a good example is swarm which studies the collective behavior of decentralized, self-organized systems \cite{b12}. These systems consist of simple agents (robots, in this case) that interact locally with each other to achieve a global objective, what can be taken as a good start to design a parallel tuning algorithm. They explore the parameter space collectively, each testing different sets of parameters in parallel, and this distributed approach helps cover a larger search space more quickly. Robots communicate their findings to the group, allowing the swarm to converge on the optimal parameters more efficiently -- results from individual evaluations can be aggregated to assess the overall performance, guiding the swarm towards better parameter configurations. As in the Particle Swarm Optimization (PSO) algorithm \cite{b13}, a popular algorithm used in swarm intelligence where each robot represents a particle in the swarm. The particles adjust their positions in the parameter space based on their own experience and the experiences of their neighbors, or in Genetic Algorithms (GA), what is yet an another approach where robots represent individuals in a population. They explore the parameter space through processes analogous to natural selection, crossover, and mutation. 

Such an approach or parallel exploration significantly reduces the time required to find optimal parameters compared to sequential tuning methods, and  scales well with the number of robots, allowing for efficient optimization, by exploring a diverse set of solutions simultaneously, to increase  the likelihood of finding effective parameters. In addition, the decentralized nature of swarm intelligence ensures the tuning is not impeded by a single fault or a single faulty measurement. 

Usually, tuning of MAV controllers takes days to achieve satisfactory results, both in terms of robustness, and performance. The literature reports numerous methods to tuning general controllers, also in application to MAVs, referred to below. The first approach which can be listed is an application of  fuzzy logic approach to maintain self-tuning capabilities \cite{b14}, where a self-tuning PID controller for path tracking by MAVs is used. The obtained performance is comparable to traditional PID controllers, taking scenarios with uncertainties and disturbances into account. A fuzzy logic enables the system to  be adapted to changing conditions, however, it required an expert knowledge about the model of the plant, and a difficult trial-and-error training session to properly set tuning mechanisms of the gains of the controller. 

The other reported approach is gain scheduling \cite{b15}, where adaptive P/PI controller is designed both in state or observer-based output feedbacks, to handle different  conditions effecti\-vely. However, this approach is computationally intensive due to the need for real-time adaptation to have a variety of operating points covered by sets of gains  by means of experiments (with multiple repetitions  to increase robustness against single faulty measurement), or using the available model (in this case, the approach  depends strongly on the accuracy of the underlying dynamic model), thus implementing gain scheduling in a MAV, faces many practical challenges.

Furthermore, a  classical adaptive pole-placement technique \cite{b16} to ensure both stability and performance of the control system by placing poles of its model in desired locations. However, this technique requires the MAV's dynamic model to be identified using data from flight tests/simulations for further tuning of the PID gains and adaptation in terms of the poles. Despite just a shortlist of parameters to be identified, such as inertia, damping ratio and motor dynamics during operation of the MAV, the method relies, again, strongly on the model.

Another approach is a deep-learning exercise using Flightmare or the use of artificial neural networks \cite{b17}. Unlike the other simulators, Flightmare offers the users the possibility to alter the underlying physical model of a MAV, and, subsequently to use artificial neural networks in  tasks such as MAV control using deep reinforcement learning. Clearly, the computational cost during training and inference of neural networks, both in training and evaluating sessions are usually time-consuming, affect real-time simulation performance, and require significant memory and processing power, limiting  scalability of the method, see \cite{referees1}.

All hitherto listed approaches share the same drawback -- they need mathematical models to run, and  are greatly time consuming. The main contribution of this paper is presenting a rapid parallel model-free auto-tuning method for swarms of MAVs (or robots), with low number of tuned parameters. 

The paper is structured as follows: Section II presents the experimental hardware platform, and is related to the outline of the controller, to enable the reader to clearly identify the gains to be tuned in parallel, Section III describes the parallel tuning procedure, Section IV provides the experimental results, and the concluding words are in Section V. 

\section{Experimental hardware platform \& controller}
\label{sec:hardware}

\begin{figure}[htbp]
\centerline{\includegraphics[width=0.45\textwidth]{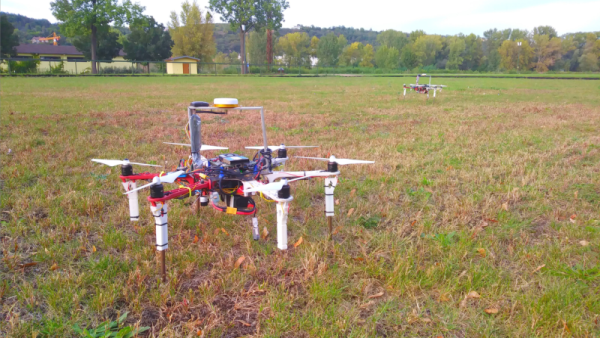}}
\caption{Hardware platforms (a pair of MAVs)}
\label{MAV_rysunek}
\end{figure}

The experimental platform, i.e.~hexarotor MAV, is built of the widely available equipment, with large potential to be mirrored by the research of the other teams. The MAV uses DJI F550 frame and  DJI E310 propulsion system. The PixHawk flight controller provides basic flight functionality, and is connected to the on-board PC \cite{MRS1,MRS2}, which sensors are used to obtain a simple information concerning position, velocity and orientation estimates in a global frame, whereas additional fusion of Garmin Lidar Lite-v3 is used for estimation of the height above the ground, including the support for cost function evaluation purposes. All the calculations are carried out on the Intel NUC-i7 (see Fig.~\ref{MAV_rysunek}).

Special attention is paid in the paper to the altitude control loop, which receives a feedback signal from  an embedded stabilizer and the error signal calculated on the basis of the desired attitude $\boldsymbol{R}_D$ and thrust $T_D$. The other loop is formed by a nonlinear $SO(3)$ state feedback controller, which receives: desired position $\boldsymbol{r}_D$ as a command, and also speed $\dot{\boldsymbol{r}}_D$, and acceleration $\ddot{\boldsymbol{r}}_D$ in a world frame. The attitude loop is  responsible for keeping the desired attitude $\boldsymbol{R}\left(\phi, \theta, \psi\right) \in SO(3)$ profile, with Euler angles $\phi$, $\theta$, $\psi$ as yaw, pitch and roll. The control signal has a form of the desired motor speed. 

The $SO(3)$ controller uses the knowledge about the model (time indices omitted) \cite{b1}
\begin{equation}
  \begin{split}
    \dot{\boldsymbol{r}} &= \boldsymbol{v}\,,\\
    m_{\mathrm{MAV}}\dot{\boldsymbol{v}} &= f \boldsymbol{R} \boldsymbol{e}_z + m_{\mathrm{MAV}}g\boldsymbol{e}_z\,,\\
    \dot{\boldsymbol{R}}&=\boldsymbol{R}\hat{\boldsymbol{\Omega}}\,,\\
    \boldsymbol{J}\dot{\boldsymbol{\Omega}} + \boldsymbol{\Omega} \times \boldsymbol{J}\boldsymbol{\Omega} &= \boldsymbol{M}\,,
  \end{split}
\end{equation}
with position $\boldsymbol{r}~=~\left[x, y, z\right]^T$, orientation $\boldsymbol{R}\left(\phi, \theta, \psi\right)$ in the world frame. The gravity force is denoted by $g \in \mathbb{R}$, and the total thrust as $f \in \mathbb{R}$, with the mass of MAV denoted as $m_{\mathrm{MAV}} \in \mathbb{R}$.
The angular velocity (body frame $\{\boldsymbol{b}_1, \boldsymbol{b}_2, \boldsymbol{b}_3\}$) is denoted by $\boldsymbol{\Omega} \in \mathbb{R}^3$ and $\boldsymbol{J} \in \mathbb{R}^{3 \times 3}$ is the  matrix of inertia.
The total moment of MAV is $\boldsymbol{M} = \left[M_1, M_2, M_3\right]^T$. 

The control law, for which the gains are to be tuned is based on  \cite{lee:so3} with a total thrust force $f \in \mathbb{R}$ as the input, and total moment $\boldsymbol{M} \in \mathbb{R}^3$ given by
\begin{align}
  \boldsymbol{M} = &-k_R\boldsymbol{e}_R-k_\Omega \boldsymbol{e}_\Omega + \boldsymbol{\Omega}\times \boldsymbol{J} \boldsymbol{\Omega} +\\
               &- \boldsymbol{J}\left(\hat{\boldsymbol{\Omega}}\boldsymbol{R}^T\boldsymbol{R}_c\boldsymbol{\Omega}_c-\boldsymbol{R}^T\boldsymbol{R}_c\dot{\boldsymbol{\Omega}}_c\right)\nonumber\,,\\
  f = &-(-k_x\boldsymbol{e}_r - k_{ib}\boldsymbol{R}\int\limits_0^tR(\tau)^T\boldsymbol{e}_rd\tau +\\
      &- k_{iw}\int\limits_0^t\boldsymbol{e}_rd\tau - k_v\boldsymbol{e}_v - m_{\mathrm{MAV}}g\boldsymbol{e}_3 + m_{\mathrm{MAV}}\ddot{\boldsymbol{x}}_d)\boldsymbol{R}\boldsymbol{e}_3\nonumber\,,
\end{align}
where $\ddot{\boldsymbol{x}}_d$ is the desired acceleration and $k_P = k_x$, $k_D = k_v$ as the to-be-tuned gains. The rotation, position and velocity errors are denoted by $e$ with  appropriate subscripts, all forming the final force $f$ and orientation $\boldsymbol{R}_C$. The symbol $k_R$ denotes a diagonal matrix standing by the rotation error $\boldsymbol{e}_R$, a diagonal matrix $k_\Omega$ standing by the angular velocity error $\boldsymbol{e}_\Omega$, and $\boldsymbol{e}_r$, $\boldsymbol{e}_v$, denote position and velocity errors, respectively. 

The control loops take advantage of the MPC (model predictive control) tracker \cite{MRS3,b2} to calculate control errors in prediction horizons, with $\boldsymbol{e}~=~\boldsymbol{x} - \boldsymbol{\hat{x}}$, and $\boldsymbol{\hat{x}}$ as the reference primitive. The MPC-related problem is solved at all time instants
\begin{align}
  & \min_{\underline{u}_{\,t}, \underline{x}_{\,t}}
  & & \mathrm{f}\left(\underline{u}_{\,t}, \underline{x}_{\,t}\right) = \frac{1}{2}\sum_{i=1}^{m-1}\left(\underline{e}^T_{\,i}\boldsymbol{Q}\underline{e}_{\,i} + \underline{u}_{\,i}^T\boldsymbol{P}\underline{u}_{\,i}\right)\nonumber
\end{align}\vspace*{-5mm}\begin{align*}
  \text{s.t.}~ \underline{x}_{\,t+1} &= \boldsymbol{A}\underline{x}_{\,t} + \boldsymbol{B}\underline{u}_{\,t}, &\forall t &\in \{0, \hdots, m-1\}\label{eq:mpc_model}\,,\\
  \underline{x}_{\,t} &\leq \underline{x}_{\,t}^{\mathrm{MAX}}, &\forall t &\in \{1, \hdots, m\}\,,\\
  \underline{x}_{\,t} &\geq \underline{x}_{\,t}^{\mathrm{MIN}}, &\forall t &\in \{1, \hdots, m\}\,,
\end{align*}
with the convex cost function in comprising penalties for control errors and input efforts in a time horizon of $m \in \mathbb{Z}_+$ samples, with $\boldsymbol{Q}\geq0$, $\boldsymbol{P}\geq0$, and equality constraints related to the model, and inequality constraints related to the admissible limits imposed on the variables. In the formulation presented, the feasible trajectory is smoothed until there are no violations of the control input constraints, not taken explicitly into account in the above-formulated problem. 

\section{Tuning procedure}

The method allows an even number of $N$  identical MAVs to be tuned at a time, while sending the updated gains to this swarm in real time. Two cases can be considered here: either when MAVs are used to average between cost function values for identical gains over the platforms, or to increase the reliability, while in the other configuration, is related to such a distribution of controller gains, as to increase the speed of tuning, by sharing different gain configurations to cover the search space, and clearly denotes a larger swarm is needed here. The first case is of paramount importance in windy environments or when using a low-cost equipment which provides measurements of poor quality. Obviously, all the measurements are low-pass filtered to reject possible noise. 

Let us consider the case of $N=2$ MAVs taking part in the experiment, with (M) as the master one and (S) as the slave MAV. It is assumed that the on-board computer of (M) carries the optimization having a 3-state state machine, whereas (S) has only 2 first states in its state machine. 

The states are as follows:
\begin{itemize}

\item IDLE, corresponding to hovering for fixed (or newly updated) gains, 
\item FLYING, when a single reference primitive is published and the cost function is collected on a MAV, to obtain the information about the performance,
\item OPTIMIZE, when a single optimization step is performed, and the updated gains are published to each MAV, and they are subsequently moved to the FLYING state. 
\end{itemize}

The MAVs are synchronized using semaphores, and a MAV is IDLE whenever it needs to wait for the end of the FLYING state in the other MAVs, or when the publication of gains by (M) has not finished yet (or the experiment has not started yet). The final tuning time clearly results from $K$, see below. 

The method is iterative-based and stores the information concerning tracking performance by adding increments to the total cost function $J_{j}$ of $j$-th MAV at sampling instants, equally spaced in time. The main iteration is run the number of times referring to the tolerance of the solution on the (M) MAV, based on the expected reduction of the initial interval where the optimal gains are expected to be. To sum up, the single step of the tuning procedure (collection of the cost function at each MAV -- (S) or (M)) is as follows: 
\begin{itemize}
\item send the reference primitive to a swarm, calculate the cost functions $J_{j}$ over a time horizon corresponding to the length of the reference primitive for the $j$-th MAV (carried out for all $N$ MAVs, $j=1,\,\ldots, N$)
\begin{equation}J^{(i)}_j = J^{(i-1)}_j + \Delta{}J^{(i)}_j\,,\end{equation}
where $\Delta{}J^{(i)}_j$ is a selected performance index to mirror the designer choice;
\item wait for all the MAVs to return to IDLE state, 

\item perform a zero-order optimization step by comparing cost function values (averaged, or not), share the updated gains, wait until the transients decay, and proceed back to publication of the reference primitive. 
\end{itemize}

The cost function $J_{j}$  mentioned above allows the user to mimic their expectations towards the control performance, by including various penalizing terms, and at the same time, offering versatility of the approach. Here, an absolute  tracking altitude error of the $j$-th MAV is used. 

The experiment-based tuning process is fully deterministic in running length (no. of steps $K$ for every parameter in a single bootstrap), and for a given zero-order method its time consumption/number of necessary iterations can be easily calculated (expressed in the number of the necessary experiments). Here, we use the EQL (equal-division) method, see \cite{b3} for details, which takes on two intermediate gain configurations at a time within the range where the optimum is sought, and afterwards, the procedure is repeated. 

Let the initial range for a given gain which is being tuned, be defined as $p\in\mathscr{R}^{(0)}=\left[p^{(0^-)},\,p^{(0^+)}\right]$, and the procedure related to the reduction of the range for a single optimized parameter $p$ is as below:
\begin{itemize}
\item  obtain the number of necessary cost function evaluations as to  reduce $\mathscr{R}^{(K)}$) below ($\hat{p}^\ast$ is the estimated optimal parameter, and $p^\ast$ is the true optimal value, unknown)
\begin{equation}|p^\ast-\hat{p}^\ast|\leq\epsilon(p^{(0^+)}-p^{(0^-)})\,,\end{equation}
\item for $k=1,\,\ldots,\,K$:
    \begin{itemize}
    \item[1)] collect cost function values for a pair of points  $\hat{p}^{(k^-)}$, $\hat{p}^{(k^+)}$ ($\hat{p}^{(k^-)}<\hat{p}^{(k^+)}$,     $\left\{\hat{p}^{(k^-)},\,\hat{p}^{(k^+)}\right\}\in \mathscr{R}^{(k-1)}$) from the current range $\mathscr{R}^{(k-1)}$;

    \item[2)] update  $\mathscr{R}^{(k)}$ with (only $p$ is changed, the remaining gains are held constant):
        \begin{itemize}
        \item[a)] for $J(\hat{p}^{(k^-)})<J(\hat{p}^{(k^+)})$, $p^{(k+1)}\in\mathscr{R}^{(k)}=\left[p^{(k-1^-)},\,\hat{p}^{(k^+)}\right]$;
        \item[b)] for $J(\hat{p}^{(k^-)})\geq{}J(\hat{p}^{(k^+)})$, $p^{(k+1)}\in\mathscr{R}^{(k)}=\left[\hat{p}^{(k^-)},\,p^{(k-1^+)}\right]$;
        \end{itemize}
    \item[3)] put $k:=k+1$;
    \end{itemize}
    \item[$\bullet$] take  $\hat{p}^\ast = \frac{1}{2}\,(p^{(K^+)}+p^{(K^-)})$.
\end{itemize}

Since the point $p^{(1^-)}$ is in the new range $\mathscr{R}^{(1)}$, the value of  $J(x^{(1^-)})$ is already known from the previous iteration (a potential reduction of the computational burden of the method when no averaging takes place in the case of $N$ MAVs), then  $p^{(1^-)}$ should be chosen to be at
$p^{(2^+)}$. In such a case, it is necessary to evaluate $J$ at a single point only, i.e.,~$p^{(2^-)}$.

For a controller comprising proportional and derivative terms as in the referred case and with a single MAV:
\begin{itemize} %
\item [0)] provide feasible ranges for $k_P$ and $k_D$;

\item [1)] set the initial value of $k_D^{(0)}$;

\item[2)] using a sequence of EQL iterations with the tolerance $\epsilon$ and $k = 0$, use the bootstrapping approach (put $k_D^{(k+1)} = k_D^{(k)}$, and optimize one parameter at a time, as in shoe lacing):

\begin{itemize}

\item [2a)]
start with initial range for $k_D$ and fixed $k_D^{(k+1)}$,  find by means of EQL the optimal ${\hat{k}_P^{\ast}}\,^{(k+1)}$, and proceed to the step 2b;

\item [2b)] start with initial range for $k_D$ and fixed $k_P^{(k+1)}={{\hat{k}_P}^\ast}\,^{(k+1)}$,  find by means of EQL the  optimal ${{\hat{k}_D}^\ast}\,^{(k+1)}$, and proceed to the step 2c;

\item [2c)] if the updated point $\left({{\hat{k}_P}^\ast}\,^{(k+1)},\,{{\hat{k}_D}^\ast}\,^{(k+1)}\right)$ has already been found in the past iterations, stop the algorithm (no improvement is possible anymore); otherwise, put $k:=k+1$ and proceed to the step 2a.

  \end{itemize}

\end{itemize}

Here, for every MAV, the altitude controller tuning procedure is based on 3 simple steps, namely, (S) -- slave, (M) -- master MAV:
\begin{itemize}
\item (S): store the sampled position  data for set gains of the controller,
\item (S); calculate the value of the cost function in the tuning procedure,
\item (M): update gains of the controllers and wait until the transient phase decays. 
\end{itemize}

As one can see on the basis of the tuning procedure, and on the basis of the properties of the EQL algorithm, after the first iteration, one of the pair of the intermediate points has already been considered, thus it can lead the way to reduce the time of calculations by almost $40\%$ in comparison with the case when all the steps are carried out. 

Next, the tuning procedure can be carried out in 2 ways:
\begin{itemize}
\item (AVG) by publishing the same gains to every MAV, and to average the cost function values gathered by the swarm, to reduce the uncertainty, negative environment impact, etc. (offering no reduction in tuning time), or
\item (without AVG) by publishing the gains to every MAV, where every MAV gets a different configuration in a single run of the EQL algorithm, to reduce the tuning horizon (yet for $N>2$ results can be averaged).
\end{itemize}

\section{Experimental results}

Prior the in-flight tuning experimental campaign at Cisarsky Ostrov in Prague has been carried out, initial tests have been run with in a realistic simulation based on the models of the hardware platform to obtain valuable insight into the performance of the tuning approach  using ROS Kinetic and Gazebo for a circular trajectory primitive with 2 m diameter and 1 m elevation difference. The lateral motion (in X and Y) tilts the MAV, the motion causes disturbances in the altitude control and it is of prime importance to obtain tuning of the altitude controller, to obtain good tracking in Z. By having a 3D trajectory (rather than 2D), the performance of the controller is put to a harder test because it has to compensate for disturbances in all 3 axes. 

\begin{figure}[htb]
\centerline{\includegraphics[width=0.5\textwidth]{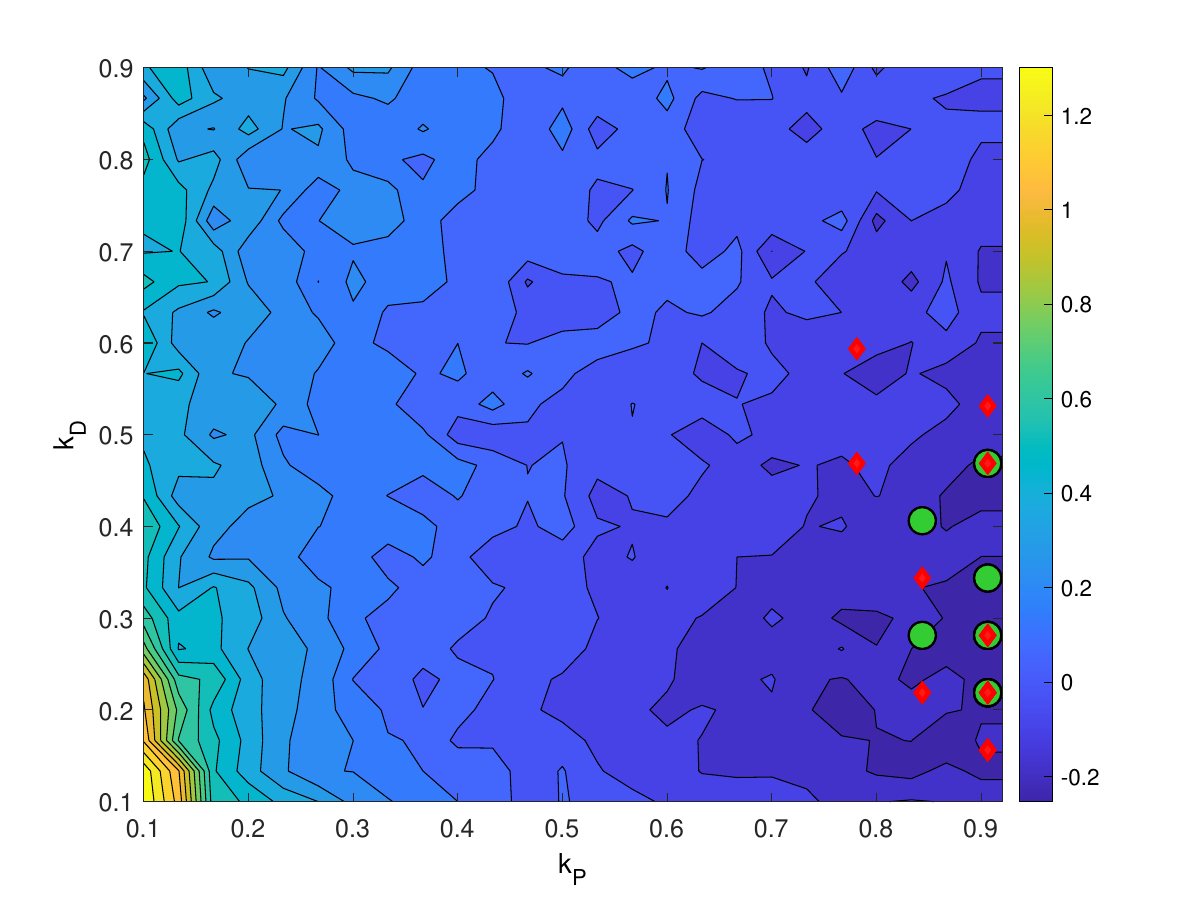}}
\vspace*{-2mm}\caption{Surface of the \underline{logarithm} of the sum of absolute tracking errors in Z averaged over 20 experiments}
\label{figsurface}
\end{figure}

The gains have been normalized, and tuned within the range $[0,\,1]$ on the basis of the simulation experiments. At first, 20 repetitions per each configuration (grid of $25\times25$ points) of the gains have been performed for a single MAV, with the \underline{logarithm} (units omitted here) of a sum of absolute tracking errors depicted in Fig.~\ref{figsurface} (the cost function surface is extremely flat, due to the action of MPC routine). 

\begin{figure}[htbp]
\centerline{\includegraphics[width=0.4\textwidth]{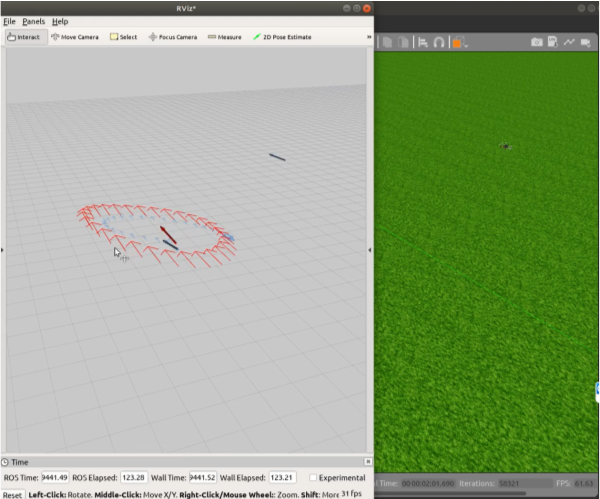}}
\caption{Gazebo simulation (RVIZ) presenting circular trajectory with the ideal/implemented thrust force vector (left) and 2 MAV models in the simulation (left)}
\label{figRVIZ}
\end{figure}

Next, a Gazebo simulation has been performed for a pair of MAVs (see~Fig.~\ref{figRVIZ}), 
following a circular trajectory and applying a parallel tuning method. The experiment has been performed 10 times, to obtain the statistics, and the tuning results were as follows ($\epsilon = 0.125$, initial value of $k_D = 0.2$, two bootstraps):
\begin{itemize}
\item without AVG
\vspace*{-5mm}\begin{eqnarray*}
k_P &=& 0.8656 \pm 0.0508\,,\\
k_D &=& 0.3094 \pm 0.1225\,,
\end{eqnarray*}

\item with AVG
\vspace*{-5mm}\begin{eqnarray*}
k_P &=& 0.9000 \pm 0.0192\,,\\
k_D &=& 0.3260 \pm 0.1051\,.
\end{eqnarray*}
\end{itemize}

\begin{figure}[htbp]
\centerline{\includegraphics[width=0.5\textwidth]{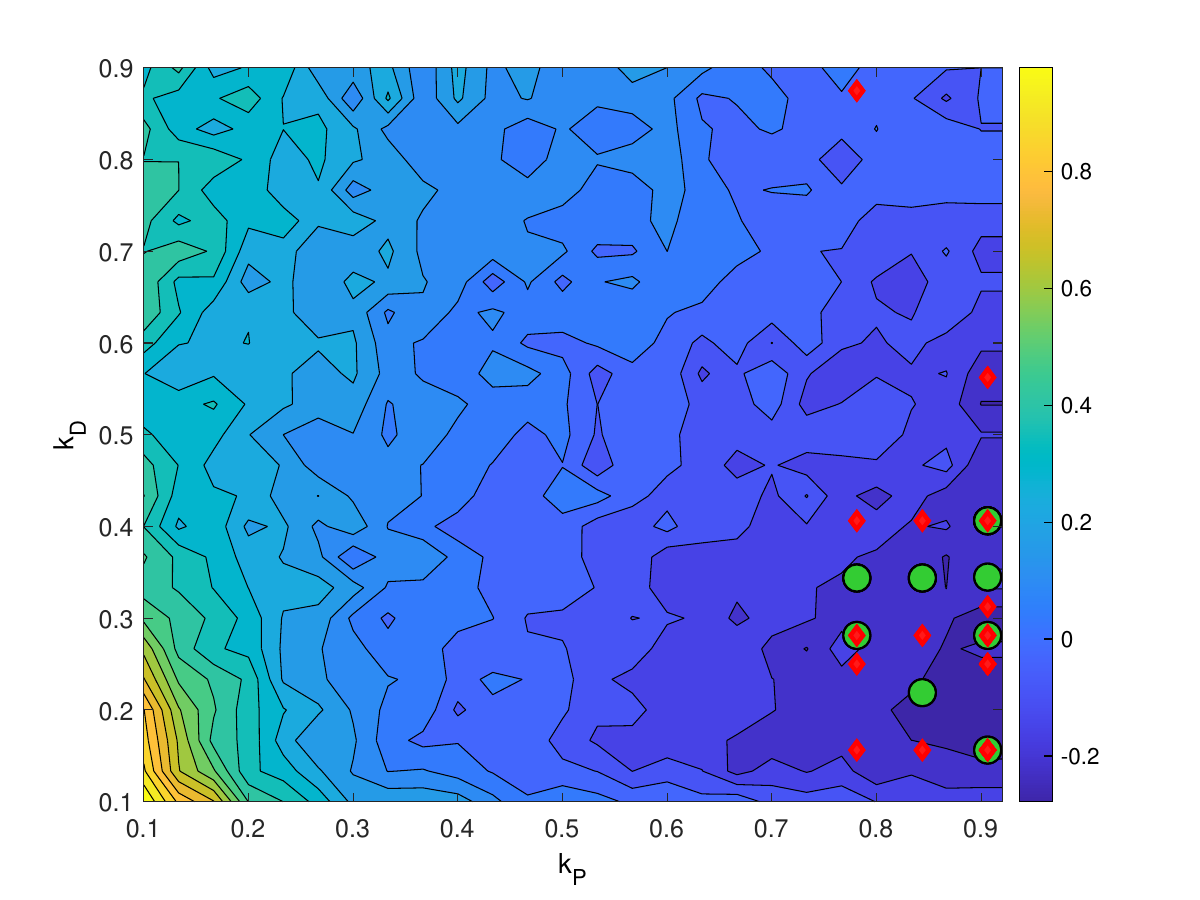}}
\vspace*{-2mm}\caption{Surface of \underline{logarithm} of the sum of absolute tracking errors in Z averaged over 20 experiments, with a payload}
\label{figsurface2}
\end{figure}

The final configuration of the gains is depicted in Fig.~\ref{figsurface2} by green dots (for AVG), and red dots (without AVG). As it can be seen, by averaging the cost function values, the final gains are produced with high repeatability in the case of AVG, what clearly indicates the way to diminish the impact of the environment on the tuning procedure. 

Each reference primitive had 8 s, with additional 1 s to decay the transient after the gains have been published, giving overall 14 cost function evaluations per single parameter in a single bootstrap, and for 2 bootstraps with 2 parameters, one gets 56 iterations, leading to a 560 s-long experiment (roughly, as per the semaphore actions and state machines switching between IDLE, FLYING and OPTIMIZE). 

In addition, the vast simulation campaign concerning impact of the load on the gains, see Fig.~\ref{figsurface2}, where the MAV is less dynamic due to an additional 900 g payload model of a Li-Po battery attached, with:
\begin{itemize}
\item without AVG
\vspace*{-6mm}\begin{eqnarray*}
k_P &=& 0.8837 \pm 0.0514\,,\\
k_D &=& 0.2712 \pm 0.1728\,,
\end{eqnarray*}

\item with AVG
\vspace*{-6mm}\begin{eqnarray*}
k_P &=& 0.8812 \pm 0.0425\,,\\
k_D &=& 0.2612 \pm 0.0904\,.
\end{eqnarray*}
\end{itemize}

The corresponding experiments on real platforms have been performed as well, omitted here for the sake of brevity, proving the gains should be adapted to payload carried, what is of prime importance in pick-and-place tasks. 

Finally, two experimental runs have been performed at Cisarsky Ostrov, with a pair of MAVs in real-world environment. Each UAV was localized using standard GPS, and the feedback information concerning position and orientation has been obtained from the MRS UAV  system \cite{MRS3} using onboard sensory data. The information to collect the cost function values has been obtained on the basis of the downward-oriented Garmin Lidar Lite v3 sensor, for accurate measurement concerning the altitude, and prove the versatility of the approach. 

\begin{figure*}[htb]
\centerline{\includegraphics[width=0.6\textwidth]{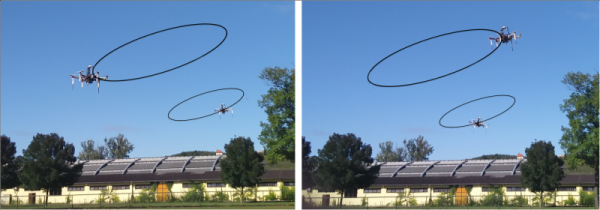}}
\caption{Circular reference trajectories}
\label{figtraj}
\end{figure*}

\begin{figure*}[htb]
\centerline{\includegraphics[width=0.95\textwidth]{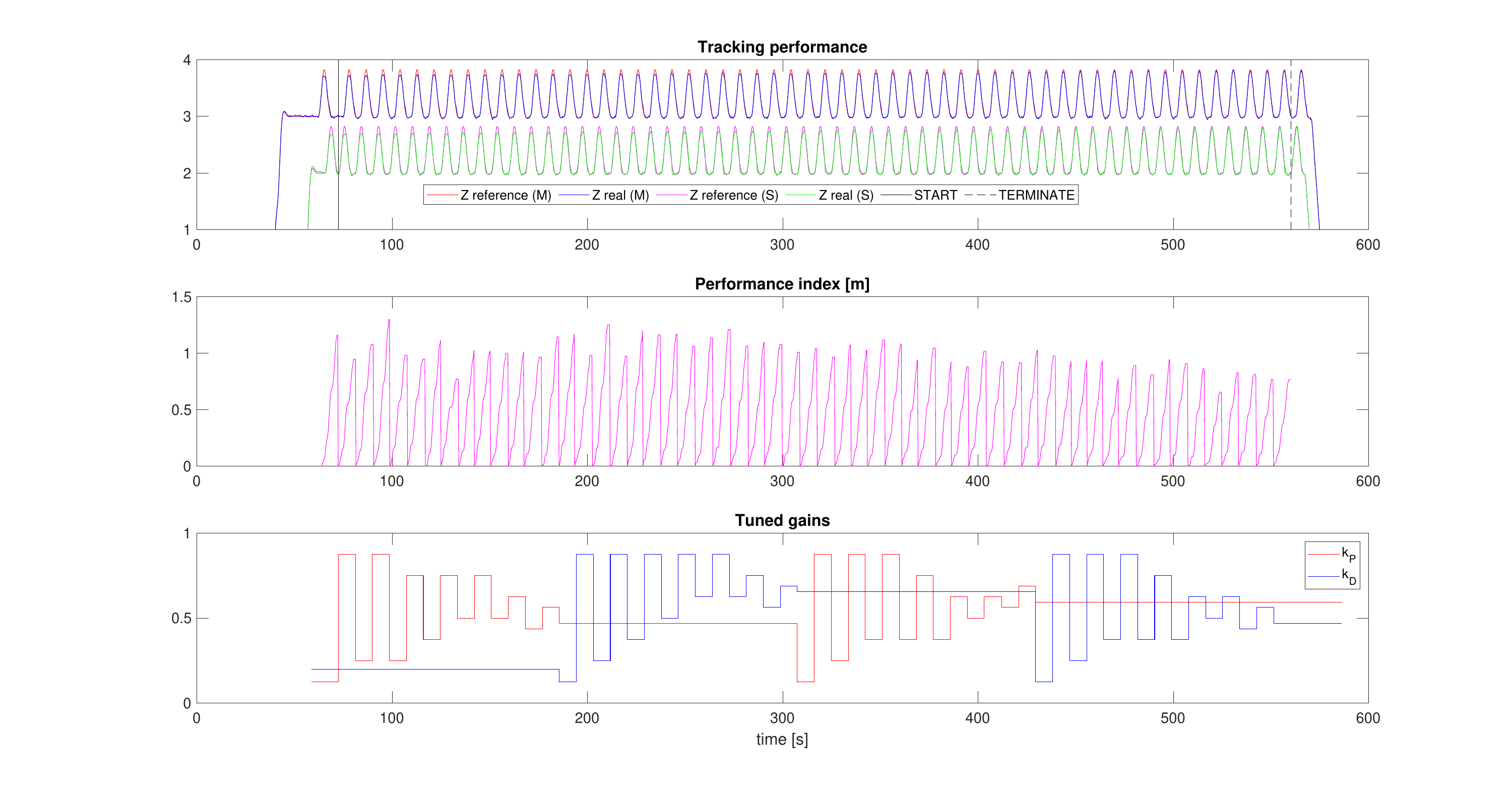}}
\vspace*{-8mm}\caption{Tuning procedure data for the case with AVG ((S) lowered in the Figure by 1 m), tracking performance in [m]}
\label{figA}
\end{figure*}

The reference trajectories have been depicted for a pair of MAVs in Fig.~\ref{figtraj}, where it can be seen they have been placed farther away from one another. 
Fig.~\ref{figA} depicts the data from the real outdoor experiment (AVG), and as can be seen, the length of the tuning procedure is longer, gains published to $N=2$ MAVs, however, leading to increase in accuracy. 

\begin{figure*}[htbp]
\centerline{\includegraphics[width=0.95\textwidth]{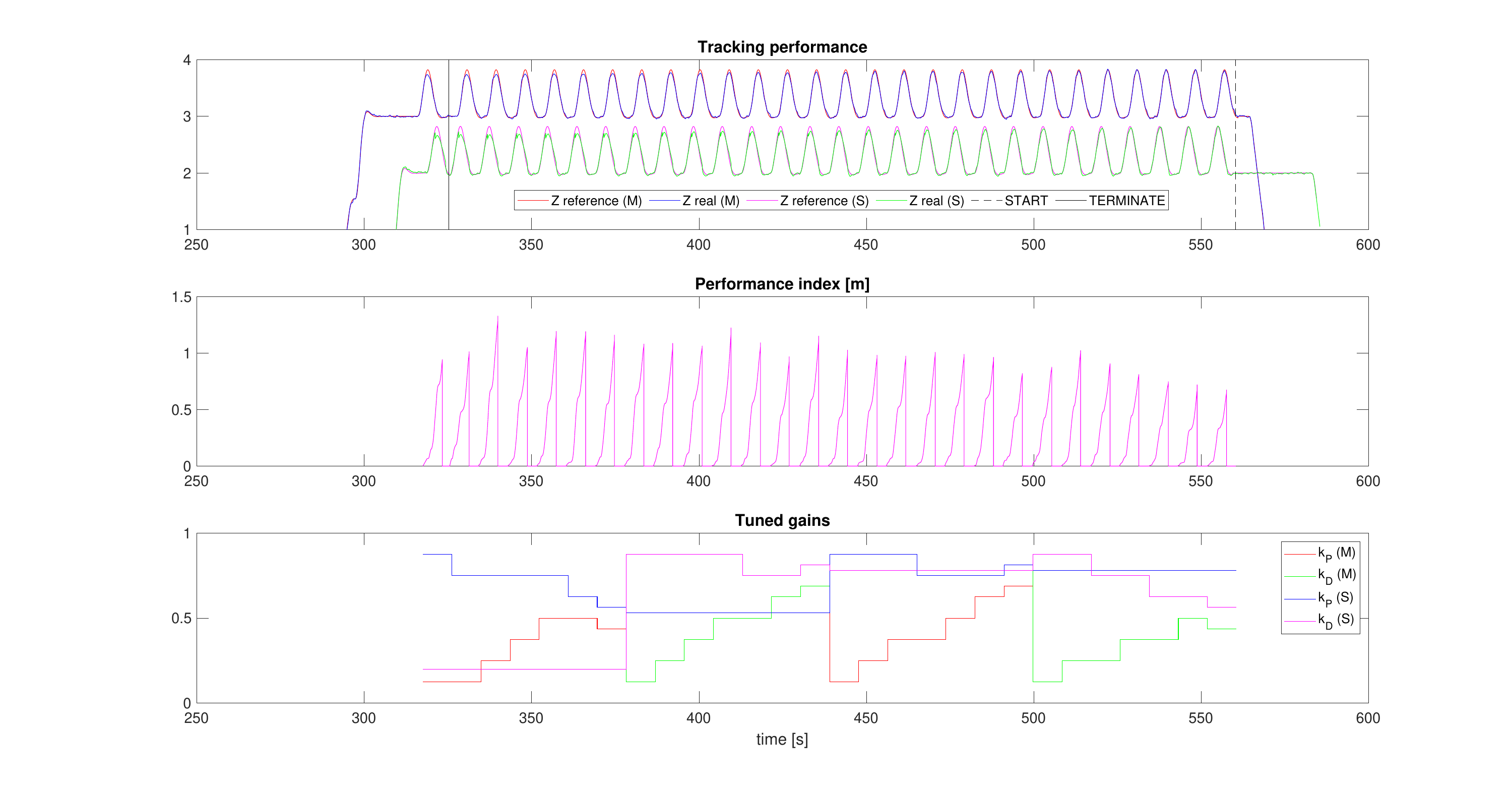}}
\vspace*{-8mm}\caption{Tuning procedure data for the case without AVG ((S) artificially lowered in the Figure by 1 m ), tracking performance axis in [m]}
\label{figwA}
\end{figure*}

On the contrary, Fig.~\ref{figwA} presents a much faster tuning (no AVG) since the gains are shared between each configuration for a single MAV only. The lowermost plot presents gain profiles, which are much different when compared its counterpart from Fig.~\ref{figA}.

The final gains have been 
\begin{itemize}
\item without AVG
\vspace*{-6mm}\begin{eqnarray*}
k_P &=& 0.5938\,,\\
k_D &=& 0.4688\,,
\end{eqnarray*}

\item with AVG
\vspace*{-6mm}\begin{eqnarray*}
k_P &=& 0.6438\,,\\
k_D &=& 0.5625\,.
\end{eqnarray*}
\end{itemize}

\section{Summary}

In the paper, we demonstrated a simple, computationally efficient, scalable and reliable method to tune swarms of MAVs in parallel by using appropriate way of distributing updated gains. The method has a huge potential to be used whenever days of trial-and-error approaches of tuning prove to be void (deterministic running time). In addition, it allows the tuning to be performed  in stages, replacing batteries, fixing hardware, and executing the remaining parts of the procedure. 
Potentially, the method leads to obtaining a bank of controller gains for various loads carried by MAVs to be addressed in a look-up table-like way by the reading of the mass estimator.

\end{document}